\documentclass[letterpaper]{article} % DO NOT CHANGE THIS
\usepackage{aaai25}  % DO NOT CHANGE THIS
\usepackage{times}  % DO NOT CHANGE THIS
\usepackage{helvet}  % DO NOT CHANGE THIS
\usepackage{courier}  % DO NOT CHANGE THIS
\usepackage[hyphens]{url}  % DO NOT CHANGE THIS
\usepackage{graphicx} % DO NOT CHANGE THIS
\usepackage{natbib}  % DO NOT CHANGE THIS AND DO NOT ADD ANY OPTIONS TO IT
\usepackage{caption} % DO NOT CHANGE THIS AND DO NOT ADD ANY OPTIONS TO IT
\usepackage{multirow}
\usepackage[linesnumbered,ruled,vlined]{algorithm2e}
\usepackage{amsmath}
\usepackage{amssymb}
\usepackage{array}
\usepackage{diagbox}
\usepackage{xcolor}
\usepackage{booktabs}
\usepackage{lipsum} % 示例文本
\usepackage{caption}
\usepackage{multirow}
\usepackage{makecell}
\usepackage{subcaption}
\usepackage{bm}
\usepackage{newfloat}
\usepackage{listings}
\DeclareCaptionStyle{ruled}{labelfont=normalfont,labelsep=colon,strut=off} % DO NOT CHANGE THIS
\title{Adaptive Dataset Quantization}
\author{
    Muquan Li,
    Dongyang Zhang\thanks{Corresponding Author.},
    Qiang Dong,
    Xiurui Xie,
    Ke Qin
}
\affiliations{
     Institute of Intelligent Computing, University of Electronic Science and Technology of China, China\\
    muquanli2023@std.uestc.edu.cn, \{dyzhang, dongq, xiexiurui, qinke\}@uestc.edu.cn

}

\usepackage{bibentry}
\begin{document}

\maketitle

\begin{abstract}
Contemporary deep learning, characterized by the training of cumbersome neural networks on massive datasets, confronts substantial computational hurdles. To alleviate heavy data storage burdens on limited hardware resources, numerous dataset compression methods such as dataset distillation (DD) and coreset selection have emerged to obtain a compact but informative dataset through synthesis or selection for efficient training. However, DD involves an expensive optimization procedure and exhibits limited generalization across unseen architectures, while coreset selection is limited by its low data keep ratio and reliance on heuristics, hindering its practicality and feasibility. To address these limitations, we introduce a newly versatile framework for dataset compression, namely Adaptive Dataset Quantization (ADQ). Specifically, we first identify the sub-optimal performance of naive Dataset Quantization (DQ), which relies on uniform sampling and overlooks the varying importance of each generated bin. Subsequently, we propose a novel adaptive sampling strategy through the evaluation of generated bins' representativeness score, diversity score and importance score, where the former two scores are quantified by the texture level and contrastive learning-based techniques, respectively. Extensive experiments demonstrate that our method not only exhibits superior generalization capability across different architectures, but also attains state-of-the-art results, surpassing DQ by average 3\% on various datasets.  
\end{abstract}
\begin{links}
\link{Code}{https://github.com/SLGSP/ADQ}
\end{links}

\section{Introduction}

\begin{figure}[t]
\centering
\includegraphics[width=0.90\columnwidth]{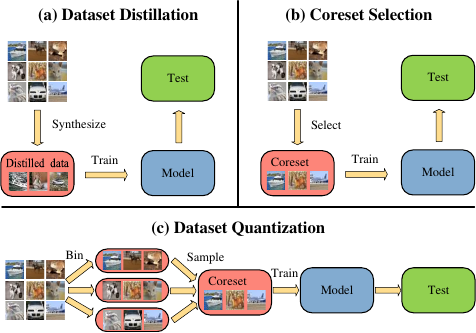}
\caption{The paradigm of three types of dataset condensation methods. The primary difference between these methods lies in the subset generating process. \textbf{(a)} Dataset Distillation synthesizes unreal dataset, \textbf{(b)} coreset selection employs one-time selection, while \textbf{(c)} dataset quantization utilizes multi-time selection as well as sampling.}
\label{fig:1}
\end{figure}

\begin{figure*}[t]
	\centering
	\begin{subfigure}{0.24\linewidth}
		\centering
		\includegraphics[width=1.0\linewidth,height=0.177\textheight]{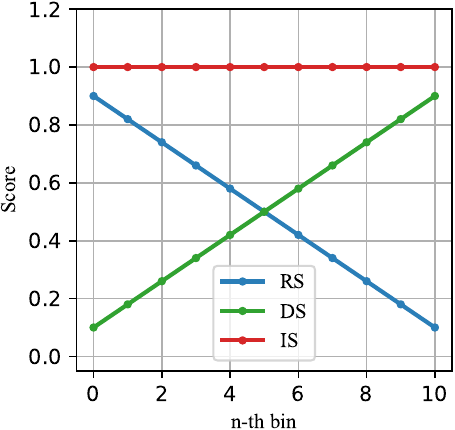}
		\caption{Ideal Condition}
		\label{fig:IC}
	\end{subfigure}
	\centering
	\begin{subfigure}{0.24\linewidth}
		\centering
		\includegraphics[width=1.0\linewidth,height=0.175\textheight]{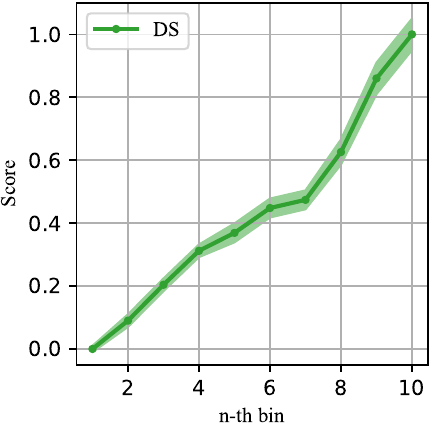}
		\caption{Representativeness Score}
		\label{fig:RS}
	\end{subfigure}	\centering
	\begin{subfigure}{0.24\linewidth}
		\centering
		\includegraphics[width=1.0\linewidth,height=0.175\textheight]{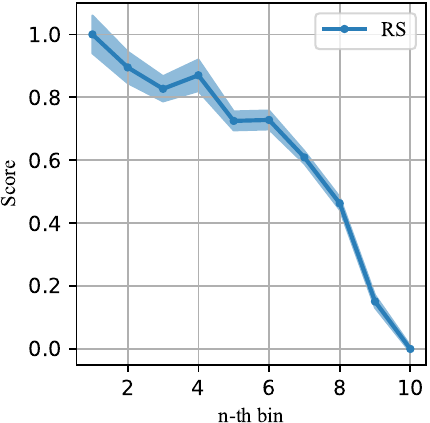}
		\caption{Diversity Score}
		\label{fig:DS}
	\end{subfigure}	\centering
	\begin{subfigure}{0.24\linewidth}
		\centering
		\includegraphics[width=1.0\linewidth,height=0.175\textheight]{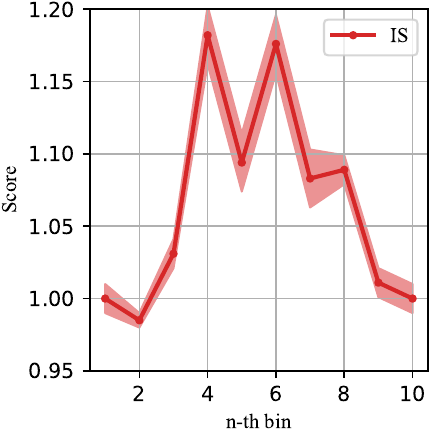}
		\caption{Importance Score}
		\label{fig:IS}
	\end{subfigure}
    \caption{The evaluation of normalized representativeness score, diversity score and importance score on CIFAR-10 \cite{krizhevsky2009learning}. \textbf{(a)} Ideal Condition allows for the best performance of DQ. \textbf{(b) (c) (d)} are representativeness score (RS), diversity score (DS) and importance score (IS) of generated bins on CIFAR-10, respectively. %The shaded areas on the curves in the figures represent the error bars.
    } 
\label{fig:2}
\end{figure*}

Deep learning has witnessed remarkable advancements recently, revolutionizing various tasks in the artificial intelligence community \cite{DBLP:conf/icml/IoffeS15}. This progress is primarily attributed to the abundance of datasets with precise labels, which serve as the foundation for training complex models. However, the expanding size of these datasets leads to increased computational costs and resource requirements. This challenge underscores the critical need for efficient dataset compression techniques \cite{DBLP:journals/pami/LeiT24}, with focus on reducing the volume of data while ensuring the consistency of training results.

In order to improve the computational efficiency, two types of techniques have made great contributions to the dataset compression, namely Dataset Distillation (DD) \cite{DBLP:conf/iclr/ZhaoMB21} and coreset selection \cite{DBLP:conf/nips/FeldmanZ20}. DD has garnered attention for its excellent performance. It aims to generate a compact but informative synthetic dataset, so that models trained with it can attain a similar or even higher level of accuracy. However, the latest optimization-oriented DD methods \cite{DBLP:conf/icml/KimKOYSJ0S22,DBLP:conf/wacv/ZhaoB23} suffer from high computational costs and poor generalization capability. Specifically, these methods employ a nested loop that alternately optimizes the distilled dataset and pre-trained model parameters\cite{DBLP:conf/cvpr/Cazenavette00EZ22b}, as well as relying on architecture-driven metrics to align the synthetic samples with the original ones \cite{DBLP:conf/iclr/ZhaoMB21,DBLP:conf/wacv/ZhaoB23}. Consequently, these limitations make it difficult to deploy DD algorithms in real-world scenario and generalize them to other model architectures. Unlike the synthesis of samples for training in DD, coreset selection aims to identify a most important subset from the training set, which has been shown to possess great cross-architecture generalization capabilities. However, as a traditional dataset compression method that employs a one-time selection strategy, its typically low data keep ratio often fails to preserve the high diversity of the whole dataset, resulting in inferior performance \cite{DBLP:conf/iccv/ZhouWGPLZYF23}. Furthermore, due to its reliance on heuristics, coreset selection cannot guarantee an optimal solution for various downstream tasks \cite{DBLP:conf/iclr/ZhaoMB21}.

To overcome the limitations of DD and coreset selection, Dataset Quantization (DQ) \cite{DBLP:conf/iccv/ZhouWGPLZYF23} is a newly proposed pipeline which first partitions the original training dataset by recursively extracting non-overlapping samples into bins based on maximizing submodular gains, and then uniformly sampled from each bin. Fig.\ref{fig:1} illustrates the main difference between DQ, DD and coreset selection. Since DQ avoids the dataset synthesis and one-time selection, it can be used for training any model architectures with high data diversity and low computational cost. The sampling strategy in DQ is based on a mathematically derived theory: the bins generated in early steps have a better representativeness of the entire dataset, while the latter bins demonstrate greater diversity. However, the naive DQ does not thoroughly analyze the uneven variations of bins' representiveness and diversity, and overlooks the varying importance of each bin, which in turn impairs the performance.

In this paper, we take a further step based on DQ, through quantitatively evaluating the importance of generated bins and introduce a novel Adaptive Dataset Quantization (ADQ). Specifically, we begin by assessing each bin through three metrics: the Representativeness Score (RS), the Diversity Score (DS), and the Importance Score (IS), which is a composite of RS and DS, corresponding to the theory in DQ. By integrating this theory into sampling strategy, we observe that DQ performs optimally merely under completely ideal condition, where the importance of each bin is equal and the trends of RS and DS resemble the blue and green curves in Fig.\ref{fig:2}(a). However, under real condition, the paucity of quantitative metrics for RS and DS precludes the appropriate estimation of IS for each bin. Therefore, to provide the evidence for precise sampling in DQ, we define three scores as following:

\textbf{\textit{Representiveness Score.}} 
Drawing inspiration from trajectories matching \cite{DBLP:conf/cvpr/Cazenavette00EZ22b,DBLP:conf/cvpr/DuJTZ023}, we propose a texture level (TL) method to calculate the representativeness score (RS) for real image sets. 

\textbf{\textit{Diversity Score.}}
As a precise method for evaluating diversity, contrastive learning-based techniques \cite{DBLP:conf/ijcai/FangSWSWS21} have been proven efficient and cost-effective, for which we introduce to calculate the diversity score (DS) of each bin.

\textbf{\textit{Importance Score.}}
It is intuitive to utilize normalization to combine the representativeness score and diversity score, yielding the importance score (IS) for each bin. 

Different from expected ideal condition in DQ, as shown in Fig.\ref{fig:2}(b)(c)(d), the real condition of these three metrics varies unevenly. It is obvious that uniform sampling strategy in DQ neglects this uneven importance variation of generated bins. Therefore, we adaptively sample from all bins based on the IS of each bin and the amount of the data it contains. Overall, the main contributions of our work can be summarized in the following three aspects:
\begin{itemize}
\item
We elucidate the sampling limitations of the naive DQ and mathematically establish appropriate metrics for evaluating the representiveness, diversity and importance of the generated bins. 
\item
We propose Adaptive Dataset Quantization (ADQ), which samples data from each generated bin according to its importance score and the number of images, achieving efficient and lossless dataset compression.
\item
Extensive experiments on CIFAR-10, CIFAR-100 
 \cite{krizhevsky2009learning}, ImageNet-1K \cite{DBLP:journals/ijcv/RussakovskyDSKS15} and Tiny-ImageNet \cite{le2015tiny} substantiate a marked enhancement in performance over the baseline DQ by average 3\%, establishing the new state-of-the-art results.
\end{itemize}

\section{Related Works}

\subsection{Dataset Distillation} 
Dataset Distillation is first proposed in \cite{DBLP:journals/corr/abs-1811-10959}, where the distilled images are expressed as model weights and optimized by gradient-based hyperparameter tuning. Subsequently, a series of bi-level optimization-oriented works seek to minimize the surrogate models learned from both synthetic and original datasets, depending on various metrics such as the matching gradients \cite{DBLP:conf/iclr/ZhaoMB21,DBLP:conf/icml/KimKOYSJ0S22,DBLP:conf/cvpr/ZhangZLMPZDLX23}, features \cite{DBLP:conf/cvpr/WangZPZYWHBWY22}, distribution \cite{DBLP:conf/wacv/ZhaoB23,DBLP:conf/cvpr/ZhaoLQY23}, training trajectories \cite{DBLP:conf/cvpr/Cazenavette00EZ22b,DBLP:conf/icml/CuiWSH23,DBLP:conf/cvpr/DuJTZ023}, and maximum mean discrepancy \cite{DBLP:conf/aaai/0003LW0G24}. However, the synthetic data from these methods often struggle to generalize across different architectures and face significant computational challenges \cite{DBLP:conf/iccv/ZhouWGPLZYF23}. Recently, a notable work \cite{DBLP:conf/cvpr/Cazenavette00EZ23} integrates a plug-and-play module GLaD into existing DD framework to improve generalization, while the high training costs remain a concern. Besides, the uni-level optimization-oriented work \cite{DBLP:conf/cvpr/Liu0LYXWN000WG22,DBLP:conf/nips/ZhouNB22} effectively reduces calculation costs but may hinder scalability to larger data keep ratio.

\subsection{Coreset Selection} 
Coreset selection \cite{DBLP:conf/nips/FeldmanZ20,DBLP:conf/dexa/GuoZB22} focuses on selecting an important subset of the original dataset, showing remarkable potential in facilitating cross-architecture training. To evaluate the subset's importance, multiple metrics have been proposed in previous work: error \cite{DBLP:conf/iclr/TonevaSCTBG19}, geometry \cite{DBLP:conf/eccv/Agarwal0AA20}, memorization \cite{DBLP:conf/nips/FeldmanZ20}, uncertainty \cite{DBLP:conf/iclr/ColemanYMMBLLZ20}, gradient-matching \cite{DBLP:conf/icml/KillamsettySRDI21}, submodularity \cite{DBLP:conf/alt/IyerKBA21}, EL2N score \cite{DBLP:conf/nips/PaulGD21}, submodular gains \cite{DBLP:conf/alt/IyerKBA21} and contributing dimension structure \cite{DBLP:conf/aaai/WanWWWZ024}. However, its low data keep ratio leads to impaired diversity of subset \cite{DBLP:conf/iccv/ZhouWGPLZYF23}, and its dependence on heuristics hinders the optimization to downstream task \cite{DBLP:conf/iclr/ZhaoMB21}. Additionally, as an extension of coreset selection, Dataset Quantization (DQ) \cite{DBLP:conf/iccv/ZhouWGPLZYF23} improves upon the traditional one-time sampling strategy by recursively generating non-overlapping bins and performing uniform sampling across all bins. This approach enhances the paradigm of sampling by shifting from a single-selection to a multi-selection strategy, thereby maintaining an appropriate data keep ratio in the subset and making it more suitable for various downstream tasks. Nevertheless, the uniform sampling strategy neglects to quantify the importance of the generated bins, for which we chose it as the baseline to address.

\subsection{Remark} 
%In the field of dataset compression, previous methods have proposed various metrics to evaluate the representiveness and diversity of a dataset. However, most of these tend to prioritize either the representiveness \cite{DBLP:conf/alt/IyerKBA21} or diversity \cite{DBLP:conf/aaai/WanWWWZ024}, rather than integrating both aspects. Besides, the evaluation methods are often either too simple, like using $L2$-norm and cosine distance \cite{DBLP:conf/wsdm/CeccarelloPP18} to assess diversity, or overly complex, like employing a pre-train model to obtain information \cite{DBLP:journals/csur/LiCWMTTL17}. Primarily, these methods rely on one-time evaluation of the entire dataset, which leads to limited precision. In contrast, we propose using texture level and contrastive learning-based method to evaluate these metrics for each generated subset, thereby achieving high precision while maintaining low computational requirements.
In the realm of dataset compression, previous studies have introduced a variety of metrics to assess the representativeness and diversity of datasets. However, the majority of these methods tend to focus on either representativeness \cite{DBLP:conf/alt/IyerKBA21} or diversity \cite{DBLP:conf/aaai/WanWWWZ024}, rather than combining both aspects. Additionally, the evaluation techniques are often either overly simplistic, such as using $L2$-norm and cosine distance \cite{DBLP:conf/wsdm/CeccarelloPP18} to gauge diversity, or excessively comrplex, like utilizing a pre-trained model to derive insights \cite{DBLP:journals/csur/LiCWMTTL17}. Typically, these methods depend on a one-time evaluation of the entire dataset, resulting in limited precision. In contrast, we propose a method that employs texture-level analysis and contrastive learning-based techniques to evaluate these metrics for each generated subset. This approach allows us to achieve high precision with low computational demands.

\begin{figure*}[t]
\centering
\includegraphics[width=2.1\columnwidth]{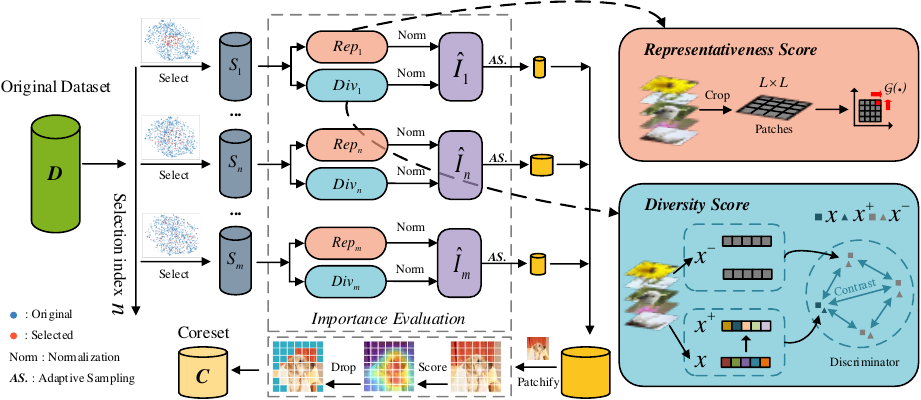}
\caption{The overall framework of the proposed Adaptive Dataset Quantization (ADQ). Following Dataset Quantization (DQ), we first divide the original dataset $\mathbf{D}$ into m non-overlapping bins $[\mathbf S_1,...\mathbf S_n,...\mathbf S_m]$. Next, an importance evaluation is conducted to calculate representativeness score, diversity score and importance score for $\mathbf{S_n}$. We then employ an adaptive sampling based on the importance score and the number of samples in $\mathbf{S_n}$ to obtain a initial compressed set. Eventually, a patch dropping and reconstruction process via MAE \cite{DBLP:conf/cvpr/HeCXLDG22} is used to drop uninformative patches, as detailed in the Appendix.} \label{fig:3}
\end{figure*}

\begin{figure}[t]
\centering
\includegraphics[width=1.0\columnwidth]{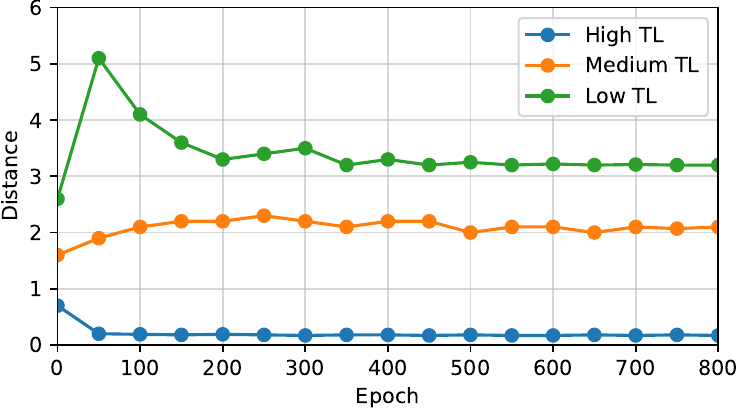}
\caption{The illustration of three types of texture level (TL) curves: High TL, Medium TL and Low TL. These curves represent the distances between the expert and the students in our improved trajectories matching.} \label{fig:4}
\end{figure}

\section{Proposed Method}

As mentioned in the Introduction section, we recognize the promising potential of DQ \cite{DBLP:conf/iccv/ZhouWGPLZYF23} and choose it as the starting point of our research. In this section, we first define the problem that DQ attempts to address. Furthermore, we analyze the clear drawbacks of naive DQ. Finally, we propose three types metrics to evaluate each bin and adaptive sampling to address these drawbacks.

\subsection{Problem Definition}
\subsubsection{GraphCut in Coreset Selection}

Let $\mathbf{D}=\left\{\left(x_{k}, y_{k}\right)\right\}_{k=1}^{M}$ represents $M$ labeled samples. By default, coreset selection involves selecting ${K}$ samples from $\mathbf{D}$ to form a coreset. The coreset is initialized as $\mathbf{S}_{1}^{1} \leftarrow \emptyset$ and updated as $\mathbf{S}_{1}^{k} \leftarrow \mathbf{S}_{1}^{k-1} \cup x_{k}$. Note that $\forall p\in\mathbf{D}$, $f(p)\in\mathbb{R}^{m\times1}$, $\mathbf{S}_{n}^{k}$ represents the first $k$ samples of the $n$-th bin and $x_{k}$ is the $k$-th selected sample, $\mathbf{S}_{1}^{k-1}$ denotes the set of selected samples, $\mathbf{D}\backslash \mathbf{S}_{1}^{k-1}$ is the remaining set and $f(\cdot)$ is the feature extractor. In GraphCut (GC) \cite{DBLP:conf/alt/IyerKBA21}, samples are selected by maximizing submodular gains $P (x_{k})$ in the feature space, defined as follows,
{\scriptsize
\begin{equation}
P\left(x_{k}\right)=\sum_{p \in \mathbf{S}_{1}^{k-1}} \underbrace{\left\|f(p)-f\left(x_{k}\right)\right\|_{2}^{2}}_{C_{1}\left(x_{k}\right)}-\sum_{p \in \mathrm{D} \backslash \mathbf{S}_{1}^{k-1}} \underbrace{\left\|f(p)-f\left(x_{k}\right)\right\|_{2}^{2}}_{C_{2}\left(x_{k}\right)}.
\label{eqn:1}
\end{equation}
}
\subsubsection{Dataset Quantization}

Almost all coreset selection methods use a heuristic metric to select samples similar to GC, making it difficult to avoid selecting samples with similar performances according to the metric. To address this selection bias, Dataset Quantization \cite{DBLP:conf/iccv/ZhouWGPLZYF23} propose a new framework consisting of three steps: bin generation, bin sampling, and pixel quantization. In detail, DQ first partitions the dataset into several non-overlapping bins. Given a dataset $\mathbf{D}$, small informative bins are recursively sampled from $\mathbf{D}$ with a predefined bin size $K$. Each bin is selected by maximizing the submodular gain described in Eqn.\ref{eqn:1}, resulting in a set of small bins $[\mathbf{S_1}, \ldots, \mathbf{S_n}, \ldots, \mathbf{S_m}]$. The selection of the $k$-th sample in the $n$-th bin is formulated as follows,
{\small
\begin{equation}
x_k \leftarrow \arg\max \left( \sum_{p \in \mathbf{S_{n}^{k-1}}} C_1(x_k) - \sum_{p \in \mathbf{D} \setminus \mathbf{S_1} \cup \cdots \cup \mathbf{S_{n}^{k-1}}} C_2(x_k) \right),
\label{eqn:2}
\end{equation}
}where $C_{1}\left(x_{k}\right)$ and $C_{2}\left(x_{k}\right)$ have been defined in Eqn. \ref{eqn:1}, $\mathbf{D} \setminus (\mathbf{S_1} \cup \cdots \cup \mathbf{S_{n}^{k-1}})$ represents the remaining data in the dataset after selecting $(k-1)$ samples in $n$-th bin. 

Following this, a uniform sampler $g(\cdot, \cdot)$ is used to sample a specific portion from each bin to form the final coreset set. Additionally, inspired by reconstructing images using only some of their patches in the Masked Auto-Encoder (MAE) \cite{DBLP:conf/cvpr/HeCXLDG22}, DQ discards less important patches to reduce the number of pixels used for describing each image. The detailed patch dropping and reconstruction strategy is described in the Appendix.

\subsection{Problem Analysis}

Although DQ achieves high coverage of the overall data across different model architectures, it encounters a significant challenge. According to the derivation of average feature \cite{DBLP:conf/iccv/ZhouWGPLZYF23}, the bin generated in the earlier steps is primarily influenced by the distances within the remaining data, while the bin in the later steps is more affected by the diversity of data in the current bin. To balance representativeness and diversity, DQ employs simple uniform sampling to randomly select an equal proportion of data from each bin. However, this uniform sampling strategy performs optimally only under an ideal condition. Specifically, given that the representativeness and diversity of each bin are unknown, their importance for inclusion in the original dataset remains uncertain. If the influence of representativeness and diversity on the results does not exhibit a uniform linear variation, as shown in Fig.\ref{fig:2}(a), then uniform sampling may only achieve a spurious balance and fail to produce the best possible outcomes.

\subsection{Importance Evaluation}

Obviously, evaluating the varying importance of the sequentially generated bins is crucial for rectifying this spurious balance. To effectively illustrate the variation in importance, we quantify three metrics for each bin, as follow:

\subsubsection{Representativeness Score}

Inspired by the trajectories matching \cite{DBLP:conf/cvpr/DuJTZ023}, quantifying the representativeness of each bin can be approached by calculating the distance between each bin and the original dataset along different training trajectories. We theoretically assume the existence of an expert parameter representing the optimal training trajectory, which corresponds to the training trajectory of the original dataset. Other training trajectories are considered student parameters. The distance between the expert and student is then calculated in the parameter space to reflect the representativeness of different bins for the entire dataset. However, traditional trajectory matching methods \cite{DBLP:conf/cvpr/Cazenavette00EZ22b,DBLP:conf/cvpr/DuJTZ023} are typically optimized through backpropagation on non-real images during dataset distillation, which contrasts with DQ that operates on real images.

Addressing this limitation, we propose a straightforward yet effective technique, termed the texture level (TL) method, as an alternative to utilizing training trajectories for trajectory matching in real images. Specifically, we first crop the images in each bin into patches $P$ of size $L\times L$. Following this, we introduce a general gradient operator $\mathcal{G(\cdot)}$ to calculate the texture level $T(\cdot)$ of each bin:
\begin{equation}
T(P) = \frac{1}{L^2} \sum_{i,j \in [1, \ldots, L]} \mathcal{G}(P_{i,j}),
\label{eqn:3}
\end{equation}
where the subscript i, j denotes the pixel coordinates.

To demonstrate the matching effect of texture level, we then crop the entire original dataset into patches and calculate the texture level of each patch. These patches are divided into three equal batches, each representing a third of the dataset: the top third are classified as High Texture Level, the middle third as Medium Texture Level, and the bottom third as Low Texture Level. Next, we train the selected model on these three batches (as the students) and on the original dataset (as the expert), while calculating a type of trajectory parameter distance in each batch with original dataset. Details about trajectory parameter distance are provided in Appendix. Intuitively, we obtain three distance curves varying with training epochs. Fig.\ref{fig:4} is obtained by training ResNet-18 on the CIFAR-10 dataset. It is observed that the distance between the student model and the expert model decreases as texture level increasing. For models trained on high-level texture patches, this distance approaches zero, indicating that images with more complex textures guide the model to progress along a trajectory more similar to that of the original dataset. Therefore, we transform the calculation of the RS $Rep(\cdot)$ for each bin into the computation of its texture level $T(\cdot)$, through $Rep(\cdot) = T(\cdot)$.

\subsubsection{Diversity Score}

We introduce a contrastive learning-based method for measuring diversity \cite{DBLP:conf/ijcai/FangSWSWS21}, modeling data diversity as an instance discrimination problem. First, we introduce a discriminator $d(\cdot)$, which is a simple multi-layer perception that takes the representation from the penultimate layer and the global pooling of intermediate features as input. In each bin $\mathbf{S}_n$, a positive view $x^{+}$ is constructed for each image using random augmentation, such as rotations, flips, and color adjustments, to enhance variability, while other images in $\mathbf{S}_n$ are considered negative views $x^{-}$. The discriminator learns to distinguish different samples by pulling positive samples closer and pushing negative samples farther apart, thereby calculating data diversity through contrastive learning. We use simple cosine similarity $cos(\cdot)$ to describe the relationship between data pairs $x_1$ and $x_2$:
\begin{equation}
  {cos}(x_1, x_2, d) = \frac{\langle d(x_1), d(x_2) \rangle}{\| d(x_1) \| \cdot \| d(x_2) \|},
\label{eqn:5}
\end{equation}

Let $\tau$ be the temperature parameter of the discriminator. The diversity $Div(\cdot)$ of data for $\mathbf{S}_n$ can then be represented as follows:
{\small
\begin{equation}
\begin{aligned}
{Div}(\mathbf{S}_n) = &- \mathbb{E}_{x_i \in \mathbf{S}_n} \left[ \mathbb{E}_{x_j \in \mathbf{S}_n} \left[ \frac{\exp(cos(x_i, x_j^{-}, d) / \tau)}{\exp(cos(x_i, x_i^{+}, d) / \tau)} \right] \right] \\
= &- \frac{1}{N(x^{-})} \left[ \mathbb{E}_{x_i \in \mathbf{S}_n} \left[ \frac{\sum_j \exp(cos(x_i, x_j, d) / \tau)}{\exp(cos(x_i, x_i^{+}, d) / \tau)} \right] \right],
\end{aligned}
\label{eqn:6}
\end{equation}
}where $N(x^{-})$ refers to the amount of negative samples for each $x_i$ in $\mathbf{S}_n$.

\subsubsection{Importance Score}
After calculating the RS and DS, we normalize \cite{DBLP:conf/icml/IoffeS15} both scores separately to facilitate the evaluation of the varying importance of generated bins on the same scale:
\begin{equation}
  \hat{Rep}_n = \text{Norm}(Rep_n, \mathbb{S}_{Rep}),
\label{eqn:7}
\end{equation}
\begin{equation}
  \hat{Div}_n = \text{Norm}(Div_n, \mathbb{S}_{Div}),
\label{eqn:8}
\end{equation}
where $Rep_n$ and $Div_n$ represent the RS and DS of the $n$-th bin, $\mathbb{S}_{Rep}$ and $\mathbb{S}_{Div}$ denote the sets of all RS and DS. We then defined the IS $\hat{I}_n$ for $n$-th bin as the sum of the normalised RS and DS:
\begin{equation}
  \hat{I}_n = \hat{Rep}_n+\hat{Div}_n,
\label{eqn:9}
\end{equation}

The variations in RS, DS, and IS of bins during the bin generation process are illustrated in Fig.\ref{fig:2}(b)(c)(d). As observed, the overall importance of bins initially increases and then decreases throughout the generation process. This pattern corresponds with our analysis of the spurious balance discussed in the Problem Analysis section. To capitalize on the dynamic importance of each bin, we introduce an adaptive sampling method.

\subsection{Adaptive Sampling}

We calculate the proportion $r_{n}$ of images to be selected from each bin based on its normalized importance value $\hat{I}_n$ and the number of images $N(n)$ in the $n$-bin:
\begin{equation}
  r_{n}=\alpha \hat{I}_n+(1-\alpha) \frac{N(n)}{\sum_{n=1}^{m} N(n)},
\label{eqn:10}
\end{equation}
where $\alpha \in [0, 1]$ denotes a weighting coefficient to balance the importance and the number of images in each bin, $m$ represents the total number of generated bins. The effect of value $\alpha$ will be discussed in Ablation Study section. Eventually, the final number of images $q_{n}$ selected from each bin is determined:

\begin{equation}
  q_{n}=\left\lfloor {r}_n \times N(n)\right\rfloor,
\label{eqn:12}
\end{equation}
where $\left\lfloor \cdot \right\rfloor$ denotes the floor function, ensuring that the total number of selected images does not exceed the required number. The adaptive process after bin generation is shown in Alg.\ref{alg:1}. Following this, a process of patch dropping and reconstruction is used to remove invalid information \cite{DBLP:conf/cvpr/HeCXLDG22}, as detailed in the Appendix. The overall framework of our ADQ is illustrated in Fig.\ref{fig:3}.

\begin{algorithm}[t]
	\caption{Adaptive Dataset Quantization}\label{alg:1}
	\KwIn{m dataset bins $\mathbf S_1,...\mathbf S_n,...\mathbf S_m$. }
	 
	% \BlankLine
	\textbf{Required:} Patch size $L\times L$, temperature parameter $\tau$ of the discriminator, weighting coefficient $\alpha$.
	
	\For{\textnormal{$n=1,...,m$}}{
		$P$ with $L\times L$ $\leftarrow$ $\mathbf S_n$
  
            Calculate $Rep_n$ using Eqn.\ref{eqn:3}

            $x^+$, $x^-$ $\leftarrow$ $x$ in $\mathbf S_n$

            Calculate $Div_n$ using Eqn.\ref{eqn:5}

            $\hat{Rep}_n$ $\leftarrow$ $\text{Norm}(Rep_n)$; $\hat{Div}_n$ $\leftarrow$ $\text{Norm}(Div_n)$

            $\hat{I}_n$ $\leftarrow$ $\hat{Rep}_n+\hat{Div}_n$

            Calculate $r_n$ using Eqn.\ref{eqn:10}

            Calculate $q_n$ using Eqn.\ref{eqn:12}	

            Select randomly $q_n$ samples from $n$-th bin
	}
        \KwOut{Initial compressed dataset.} 
	
\end{algorithm}

\begin{figure*}[t]
	\centering
	\begin{subfigure}{0.24\linewidth}
		\centering
		\includegraphics[width=1.0\linewidth]{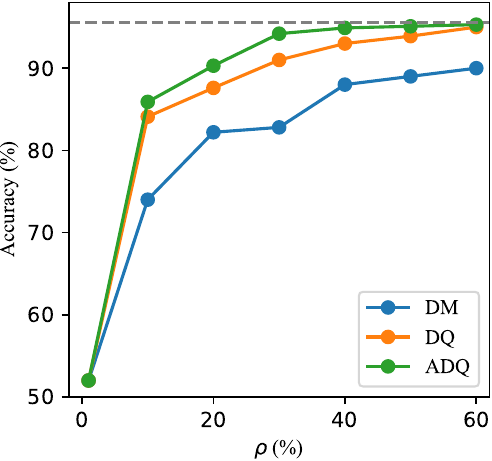}
		\caption{High Ratio}
		\label{fig:5:a}%
	\end{subfigure}
	\centering
	\begin{subfigure}{0.24\linewidth}
		\centering
		\includegraphics[width=1.0\linewidth]{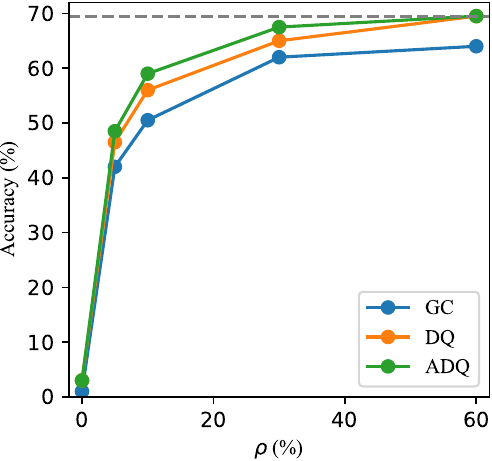}
		\caption{High Ratio}
		\label{fig:5:b}
	\end{subfigure}	\centering
	\begin{subfigure}{0.24\linewidth}
		\centering
		\includegraphics[width=1.0\linewidth]{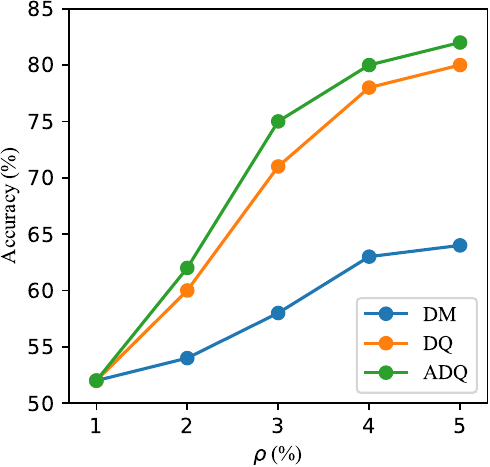}
		\caption{Low Ratio}
		\label{fig:5:c}
	\end{subfigure}	\centering
	\begin{subfigure}{0.24\linewidth}
		\centering
		\includegraphics[width=1.0\linewidth]{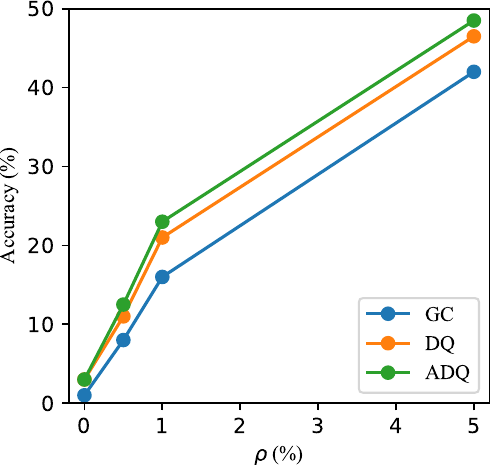}
		\caption{Low Ratio}
		\label{fig:5:d}
	\end{subfigure}
\caption{The performance of DM \cite{DBLP:conf/wacv/ZhaoB23}, DQ \cite{DBLP:conf/iccv/ZhouWGPLZYF23} and ADQ on \textbf{(a)} high data keep ratio and \textbf{(c)} low data keep ratio on CIFAR-10; and GC \cite{DBLP:conf/alt/IyerKBA21}, DQ and ADQ on \textbf{(b)} high data keep ratio and \textbf{(d)} low data keep ratio on ImageNet-1K. The dashed lines in grey in \textbf{(a)} and \textbf{(b)} indicate the results when the data keep ratio is 100\%. 
%Each reported result is the average of 5 experiments.
} 
\label{fig:5}
\end{figure*}   

\begin{table*}[t]\centering
\renewcommand\arraystretch{1.0}
% \small 
\begin{tabular}{c|cccc|cccc|cccc}
\toprule
      & \multicolumn{4}{c|}{DM}    & \multicolumn{4}{c|}{DQ}    & \multicolumn{4}{c}{ADQ}   \\ 
      \midrule
$\rho$ (\%)    & 10   & 20   & 30   & 100  & 10   & 20   & 30   & 100  & 10   & 20   & 30   & 100  \\
      \midrule
ResNet-18   & 74.0 & 82.2 & 82.8 & 95.6 & 84.1 & 87.6 & 91.0 & 95.6 & 86.2 (+2.1) & 90.4 (+2.8) & 94.2 (+3.2) & 95.6 \\
ResNet-50   & 35.0 & 36.2 & 43.9 & 95.5 & 82.7 & 88.1 & 90.8 & 95.5 & 84.7 (+2.0) & 90.7 (+2.6) & 93.7 (+2.9) & 95.5 \\
ViT   & 21.6 & 25.5 & 23.1 & 80.2 & 58.4 & 66.8 & 72.0 & 80.2 & 61.1 (+2.7) & 69.8 (+3.1) & 74.7 (+2.7) & 80.2 \\
Swin  & 25.1 & 30.1 & 27.3 & 90.3 & 69.2 & 79.1 & 84.4 & 90.3 & 73.2 (+4.0) & 82.5 (+3.4) & 88.5 (+4.1) & 90.3 \\
ConvNeXt & 41.8 & 48.3 & 47.9 & 73.0 & 52.8 & 61.8 & 64.2 & 73.0 & 55.0 (+2.2) & 64.0 (+2.2) & 68.1 (+3.9) & 73.0 \\ 
      \midrule
Average  & 39.5 & 44.5 & 45   & 86.9 & 69.4 & 76.7 & 80.5 & 86.9 & \textbf{72.0 (+2.6)} & \textbf{79.5 (+2.8)} & \textbf{83.8 (+3.3)} & 86.9 \\ \bottomrule
\end{tabular}
\caption{Comparisons of DM \cite{DBLP:conf/wacv/ZhaoB23}, DQ \cite{DBLP:conf/iccv/ZhouWGPLZYF23} and our ADQ on CIFAR-10 with different data keep ratios $\rho$. The training processes are implemented across five various architectures, with ResNet-18 used as the feature extractor to obtain distilled data. Each reported result is the average of 5 experiments.} \label{tab:1}
\end{table*}

\begin{table}[t]\centering
\renewcommand\arraystretch{0.8}
\begin{tabular}{c|ccc}
\toprule
Method & Number of runs & Error bars& GPU hours\\ 
\midrule
DM  & $5$ & $\pm 0.5$   & $91 h$       \\
DQ & $3$ & $\pm 0.4$  & $1 h$        \\
ADQ & $5$ & $\pm 0.2$ & $1 h$        \\ \bottomrule
\end{tabular}
\small \caption{Comparisons of number of runs, error bars and GPU hours for compressing dataset of DM, DQ and ADQ. }\label{tab:2}
\end{table}

\section{Experiments}

\subsection{Experimental Setup}
\subsubsection{Datasets}
Following the evaluation protocol of previous DQ \cite{DBLP:conf/iccv/ZhouWGPLZYF23}, we utilize image classification as a proxy task for evaluation and mainly assess our method on CIFAR-10 \cite{krizhevsky2009learning} and ImageNet-1K \cite{DBLP:journals/ijcv/RussakovskyDSKS15}. CIFAR-10 contains 50,000 samples for training and 10,000 samples for validation, with a resolution of $32\times32$. ImageNet-1K comprises 128,1126 samples from 1000 categories for training, with each category containing 50 images for validation.

\subsubsection{Implementation details}
Unless specified, we mainly use the ResNet-18 \cite{DBLP:conf/cvpr/HeZRS16} and Vision Transformer (ViT-base) \cite{DBLP:conf/iclr/DosovitskiyB0WZ21} models as the feature extractor for CIFAR-10 and ImageNet-1K, respectively. To assess the generalization of the compressed dataset, the training processes are implemented on several representative transformer and CNN architectures, including ResNet-18, ResNet-50 \cite{DBLP:conf/cvpr/HeZRS16}, ViT, Swin transformer \cite{DBLP:conf/iccv/LiuL00W0LG21}, ConvNeXt \cite{DBLP:conf/cvpr/0003MWFDX22} and MobilenetV2 \cite{DBLP:conf/cvpr/SandlerHZZC18}. During bin generation, the experimental procedure is consistent with those in DQ \cite{DBLP:conf/iccv/ZhouWGPLZYF23}. For comparison, we conduct training for 200 epochs on the CIFAR-10 with batch size 128, and we employ a cosine-annealed learning rate that initializes at 0.1. For ImageNet-1K, the training is in Distributed Data Parallel manner with the default scripts for different architectures. We conduct 5 experiments to average the results. For more details about the reproduction of the paper, please refer to the Appendix.

\subsection{Comparisons with Previous Methods}
Tab.\ref{tab:1} and Fig.\ref{fig:5}(a) present a comparison of our method with previous DM \cite{DBLP:conf/wacv/ZhaoB23} and DQ \cite{DBLP:conf/iccv/ZhouWGPLZYF23} on CIFAR-10 dataset. DM is the pioneering method that approaches data condensation via distribution matching. DQ is the first method to divide the full distribution into non-overlapping bins and then uniformly sampling from each bin, working as our baseline. In line with DQ, we use three data keep ratios (10\%, 20\%, and 30\%) to evaluate the performance variations, in addition to the 100\% ratio for a comprehensive comparison. The results reveal that datasets generated from DQ and ADQ retain higher performance levels when tested with new architectures during training. Notably, our ADQ consistently outperforms DQ across all five architectures, with average improvements of 2.6\%, 2.8\%, and 3.3\% at these ratios, respectively. The performance gains with higher data keep ratios are attributed to the increased number of effective samples available for calculating RS and DS, which in turn enhances the accuracy of the IS. For ImageNet-1K, we substitute DM with GraphCut (GC) \cite{DBLP:conf/alt/IyerKBA21}, and observe similar performance improvements with ADQ, as illustrated in Fig.\ref{fig:5}(b). 

Following DQ, we extend our performance comparisons to low data keep ratios to further highlight the metrics of ADQ, as depicted in Fig.\ref{fig:5}(c)(d). For lossless compression, our ADQ also achieves lossless results with only 60\% of the data, matching the performance of the current state-of-the-art dataset compression methods \cite{DBLP:conf/iccv/ZhouWGPLZYF23}. Turning the attention to the practical aspects of dataset generation, Tab.\ref{tab:2} provides a comparison of our ADQ with DM and DQ in terms of the number of runs, error bars, and GPU hours required. Our ADQ exhibits a reduction in average error bars across all experimental conditions. Notably, the computational modules we introduce for importance evaluation contribute negligible additional processing time. As a result, the time ADQ requires for dataset generation is on par with that of DQ and is a mere 1.1\% of the time needed by DM, underscoring ADQ’s efficiency.

\begin{table}[t]\centering
\renewcommand\arraystretch{0.8}
\begin{tabular}{l|cccc}
\toprule
Dataset & \multicolumn{4}{c}{CIFAR-10} \\ 
      \midrule
$\rho$ (\%)      & 10       & 20      & 30   & 100  \\ 
      \midrule
DQ      &         84.1 & 87.6 & 91.0    & 95.6    \\
+ RS     &     85.1     &   88.8      &  93.0    & 95.6    \\
+ DS     &     85.3     &  88.9       &  93.1     & 95.6   \\
+ IS (RS+DS)     &     86.2     &   90.4      &   94.2  & 95.6     \\ \bottomrule
\end{tabular}
\small \caption{Ablation study on RS, DS and IS, training on CIFAR-10. DQ presents our baseline. The increasing accuracy of results with incorporating three modules demonstrates the effectiveness of our ADQ.} \label{tab:3}
\end{table}
\begin{figure}[t]
\centering
\includegraphics[width=1.0\columnwidth, height=0.175\textheight]{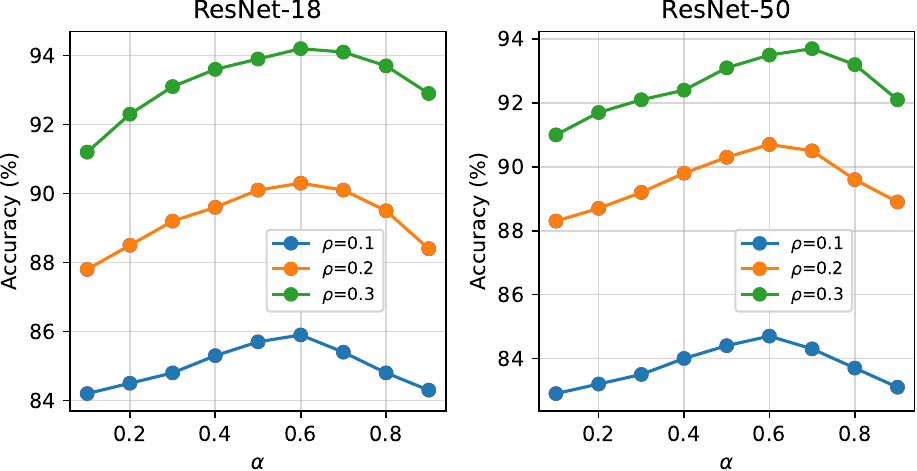}
\caption{Ablation study on the weighting coefficient $\alpha$. ResNet-18 and ResNet-50 are utilized as training models, and the experiments are implemented on three different values of data keep ratio. The average accuracy is reported.} \label{fig:6}
\end{figure}

\subsection{Ablation Study}

\subsubsection{Module Cut-off}
The ablation study begins by evaluating the contributions of the proposed three metrics of the bin: RS, DS and IS. As shown in Tab.\ref{tab:3}, DQ serves as our baseline, and its performance on CIFAR-10 is intuitively enhanced by incorporating RS, DS and IS, with averages improvements of 1.41\%, 1.53\% and 2.57\%, respectively. Note that the improvement of DS is slightly higher than that of RS across all data keep ratios, which suggests that diversity plays a more critical role than representativeness in impacting the final performance of the subset. Nevertheless, given the fluctuating conditions across different datasets and the potential rise in computational complexity due to tuning the weight ratio between representativeness and diversity (yielding only slight improvements), we maintain equal weighting for both factors when computing the importance score.

\subsubsection{Hyper-parameter analysis}
There are three hyper-parameters for ADQ: the numbers of bins $m$, the drop ratio $\theta$ and the weighting coefficient $\alpha$, where the first two parameters have been proven to give the optimal trade-off with $m=10$ and $\theta=25\%$ in DQ. Fig.\ref{fig:6} illustrates how the performance of our ADQ varies with different choices of $\alpha$. We conduct the data-keep-ratio-dependent experiments on CIFAR-10 cross two architectures, ResNet-18 and ResNet-50. As observed, accuracy initially increases and then decreases as $\alpha$ ranges from 0 to 1, reaching its peak between 0.6 and 0.7. Interestingly, the peaks of accuracy on both two datasets are shifting back (closer to 0.7). Given $\alpha$ presents the weighting of the importance score in normalized importance score (Eqn.\ref{eqn:10}), we ascribe this trend to the increased number of evaluation bases, where higher data keep ratio provides more data for assessing the importance score. As $\alpha$ approaches 0, ADQ reverts to DQ, resulting in performance that mirrors that of DQ. During the actual experiment, we adjust corresponding values of $\alpha$ according to different architectures.

\section{Conclusion}

In this paper, we introduce an Adaptive Dataset Quantization (ADQ) approach designed to address the suboptimal performance of the naive DQ method, which overlooks the differing significance of the produced bins. Specifically, we delineate three metrics for each bin: the RS, the DS, and the IS. We then employ a texture-level method and a contrastive learning-based method to compute the RS and DS, respectively. Ultimately, the IS is obtained by integrating the RS and DS, which facilitates ADQ based on the bin’s importance. Extensive experimental results confirm the efficacy of our ADQ, showing a comprehensive enhancement over the naive DQ. For future research, we intend to investigate the application of ADQ in various downstream tasks, such as object detection, image restoration.

\appendix

\section{Acknowledgments}
This work is partially supported by grants from the China Postdoctoral Science Foundation (No.2024M760357), the Postdoctoral Fellowship Program of CPSF ( No.GZB20240115), Sichuan Central-Guided Local Science and Technology Development (No.2023ZYD0165), the National Natural Science Foundation of China (NO.62176046) and Noncommunicable Chronic Diseases-National Science and Technology Major Project (No.2023ZD0501806).

\bibliography{aaai25}

\end{document}